\documentclass[12pt]{article}

\usepackage[utf8]{inputenc}

\usepackage{fullpage}

\usepackage{natbib}
\bibpunct{(}{)}{;}{a}{,}{,}

\usepackage{url}
\usepackage[bf]{caption}

\usepackage{subfig}

\usepackage{multicol}
\usepackage{multirow}

\usepackage{dcolumn}

\usepackage{amsmath}
\usepackage{amssymb}

\usepackage{boxedminipage}

\usepackage{listings}

\usepackage{minitoc}

\usepackage{ifpdf}
 
\usepackage{setspace}   

\ifpdf
\usepackage[pdftex]{graphicx}
\else
\usepackage{graphicx}
\fi

\title{Tracking Amendments to Legislation and Other Political Texts with a Novel Minimum-Edit-Distance Algorithm: \textit{DocuToads}.}

\date{}

\author{Henrik Hermansson\footnote{Post-Doctoral Researcher at the Centre for European Politics, Department of Political Science, University of Copenhagen (hajh@ifs.ku.dk).} \and James P. Cross\footnote{Lecturer in European Public Policy, School of Politics and International Relations, University College Dublin (james.cross@ucd.ie).}}

\date{}

\begin{document}

\ifpdf
\DeclareGraphicsExtensions{.pdf, .jpg, .tif}
\else
\DeclareGraphicsExtensions{.eps, .jpg}
\fi

\maketitle

\noindent
\small

\begin{abstract}

Political scientists often find themselves tracking amendments to political texts. As different actors weigh in, texts change as they are drafted and redrafted, reflecting political preferences and power. This study provides a novel solution to the problem of detecting amendments to political text based upon minimum edit distances. We demonstrate the usefulness of two language-insensitive, transparent, and efficient minimum-edit-distance algorithms suited for the task. These algorithms are capable of providing an account of the types (insertions, deletions, substitutions, and transpositions) and substantive amount of amendments made between version of texts. To illustrate the usefulness and efficiency of the approach we replicate two existing studies from the field of legislative studies. Our results demonstrate that minimum edit distance methods can produce superior measures of text amendments to hand-coded efforts in a fraction of the time and resource costs.

\end{abstract}

\thispagestyle{empty}
\clearpage

\section*{Introduction}

Political actors often engage in drafting political or legal texts to represent the policies or decision outcomes made within a political system at a given point in time. This is the case for fundamentally important documents like constitutions and international treaties, but also true for secondary legislation, party manifestos, policy statements, and even political speeches. As different actors weigh in, texts are carefully (for the most part!) drafted and redrafted until they reflect actor preferences and their relative influence over policy outcomes. Actors will insert different amendments reflecting their policy positions and what they deem to be feasible given the other actors and institutions in play.\footnote{Essentially what is meant by this is that positions and the documents that represent them are strategic in nature.} Analysing how these documents evolve, and how different actors amend documents at different points in the drafting process, can therefore tell us a lot about what actors want and how successful they are in shaping political texts to reflect these demands. 

The fact that politics is often expressed through the drafting of political texts has encouraged political scientists to track the progression of these texts in order to study a whole host of interesting questions in the field. For instance, one can study principle-agent problems in which agents are delegated with preparing a draft of a political text, and principles have incomplete control over how agents complete this task. The degree to which a text changes between a principle's version and an agent's version of a text captures not only of the power of the principal but also of the accuracy with which the agent has succeeded in predicting the preferences of the principal. Examples of this process include political speeches drafted by speechwriters and amended by the politician giving the speech, or legislative proposals drafted by committees and amended in plenary \citep{ritter2004presidential, vaughn2006conceptualizing, dille2000theprepared, schlesinger2008white}. 

Capturing amendments made to a text, especially if broken down by topic, can also be a useful indicator of extent and focus of political censorship \citep{Cross2014seen, cross2013striking, cross2014trans}. Prominent examples include the altering of websites and news articles by government agencies or editorial staff, or the detection of redaction in legislative records. One can get a strong indication of the types of issues to which censorship is being applied by focusing on what changes between an open and censored version of a text.

Finally, measuring text amendments is useful in the study of routine decision-making, when the decisions being taken are committed to text. In such cases, differences detected between documents can aids our understanding of how formalized, Weberian bureaucracies work and communicate. Document drafting processes may of course simultaneously reflect power struggles, quality control mechanisms, censorship and routine drafting work. What is clear is that the changes that occur between different versions of a text reflect real-world political and bureaucratic processes that have been the subject of academic interest for a long time. Finding an efficient, transparent and replicable method for tracking text amendment processes is thus a worthwhile undertaking.

This study introduces minimum-edit-distance algorithms to the study of the evolution of political texts. These methods have been developed in the fields of bio-informatics and computational linguistics to compare strings of DNA and speech texts respectively. The study builds upon these efforts and adapts minimum edit distances so that they become suitable for the particular aspects of the text-drafting process found in political settings. We demonstrate the usefulness of these measures by comparing their output to existing efforts to track text amendments based upon costly and time consuming hand-coded effort. We demonstrate that our approach provides a transparent and easily replicable measure of text change that is comparable across different political contexts. Our approach produces more detailed and reliable outputs that are unperturbed by human error when compared to hand-coded measurements. We also demonstrate a significant gain in efficiency, as our approach can be applied to a much larger selection of texts than would be feasible with hand coding alone. We show that our method is language-insensitive in the sense that it can trace changes between texts of the same language, irrespective of what that language is, making it uniquely suited for cross-national comparative studies. Our method can also handle different alphabets and thereby accommodate the study of a more diverse set of political systems. Finally, minimum edit distance algorithms can identify exact amendments, and the semantic context of these amendments, allowing for a much more fine-grained analysis than has previously been possible with existing automated methods in the field.

\section*{Text-as-data and the data-generating process of legislative texts}

As stated above, political science researchers have long been interested in tracking the evolution of a large diversity of political texts. Such texts include political speeches, party manifestos, legislation, court decisions, international treaties, and articles in the press. However, these different types of text are produced in very different ways, and considering \textit{how} they are produced is of vital importance when deciding upon the appropriate methods for their analysis. For instance, the manner in which a political speech is drafted and redrafted is likely to be a much more fluid and ``creative" process than the drafting and redrafting of a piece of legislation. From this it naturally follows that different methods will be more or less suitable for analysing different texts, depending on the data-generating process from which a text emerges.

Minimum edit distances are most appropriately used to compare texts where one expects parts of each text to directly correspond to one another. This is because the algorithms detects similarities between two texts using sequences of words common to both. Such a (text-as-) data structure is likeliest to emerge when political texts are drafted and redrafted, rather than created anew each time. The most obvious case of such a data generating process is in the drafting of legislation, in which each iteration of a legislative proposal will contain sections that remain un-amended and sections that are amended. Another example could be routine bureaucratic forms or communication templates which are re-used time and again with only small alterations. To demonstrate the usefulness of our method, we focus on the legislative drafting process in the European Union and German Bundestag.

While minimum edit distances can reliably show the amount and type of alterations required to change one text into another, researchers must consider whether those changes are substantively meaningful. Very often political scientists think about meaningfulness in terms of impact on society. In contrast to other written or verbal messages, legislative texts (when entering into force) have a unique impact on the real world, establishing the rights and responsibilities of states, citizens, companies etc.. Any single change in such texts may therefore have substantive consequences for those subject to the law.\footnote{The same is of course also true for international treaties, trade agreements as well as judicial rulings. Minimum edit distances may therefore provide a useful method for the study of the evolution of such texts.}

Furthermore, the legislative drafting process usually consists of definitions of concepts or of the rights and responsibilities of one party to another. Over time, and in order to reduce legal ambiguities, the structure, style, vocabulary and grammar of these definitions become subject to very strong norms and best practices within a polity.\footnote{As an example, in the European Union these are summarised here: \url{http://ec.europa.eu/governance/better_regulation/documents/legis_draft_comm_en.pdf} and here \url{http://eur-lex.europa.eu/en/techleg/index.htm}} The formalistic and precision-centered nature of legislative texts means that there are very few, if any, alternative ways of expressing the same legislative message. Precision is indeed one of the guiding principles of legislative drafting. Ambiguities may of course arise by accident, or be unavoidable, but recent research shows that they often represent a conscious attempt by the drafter to refer the interpretation to relevant courts \citep{wallace2012bringing}. Even ambiguities are thus most often deliberate.\footnote{In contrast, other political texts such as speeches are not subject to the same norms. These types of rhetorically motivated texts are instead expected to include as much linguistic variation as possible, to engage the listener or reader. Consequently, precision is also much less of a concern.}
 
The singular characteristics of legislative texts imply that political conflict in legislative bodies is almost always focused on the exact wording of laws. The institutional rules involved in the writing of laws, i.e. voting procedures and veto players, further ensures that making changes to legislative texts is difficult, meaning that spurious or non-salient changes are unlikely to be successful.\footnote{Similar voting procedures regarding the production of texts also exist outside legislatures and other representative bodies. Examples include any texts amended and adopted at general meetings of organisations, including political parties and notably their party manifestos and programs. Wikipedia entries and other texts developed in a similarly decentralised and participant-driven way could also be candidates. Finally, documents created through a process of scientific review such as the often politically controversial Intergovernmental Panel of Climate Change reports show some of these characteristics.}

For texts where `every word matters', substantively, legally, and politically, the number of words that have been edited between versions is a credible indicator of the degree of substantively important change that has been made. In our view, legislative texts provide the best example of these conditions holding true. For other types of political texts, this is not always the case and we advise users of minimum edit distance algorithms to consider carefully the question of measurement validity when applying these algorithms.\footnote{The commonly cited phrase `garbage in, garbage out' is particularly pertinent here.} If it is expected that the ordering of words in a pair of texts is important, that sections of one text will be found in another text, and finally that commonalities and differences between these texts have substantive meaning, then minimum edit distances may well be useful in examining how these texts are related. The key consideration in deciding whether or not to use our method to examine a pair of political texts is whether or not the data generating process from which a pair of texts emerge is amenable to such an analysis. Before describing our method in more detail, we provide an overview of existing approaches to tracking the evolution of political texts found in the literature.

\section*{Existing literature}

Measuring the evolution of political documents has to date proven challenging in terms of replicability and validity of the measures applied to capturing such changes, and the time and resources spent constructing such measures. Ideally, political scientists want exact measures that capture the nature and extent of changes, so that the success of individual actors in affecting outcomes can be examined. Having such measures allows one to link the (strategic) position taking of actors, the institutional context in which such positions are taken, and the final outcome that is achieved, thus providing a comprehensive account of the decision-making process. As a result, huge efforts have been made across a whole set of subfields in political science to create such measures. This literature varies significantly in the methodologies employed, with methods including close manual readings of relevant documents and large-scale quantitative analysis of political texts and speeches. Here we provide a brief overview of examples from this literature, organised by the methodology employed to capture text amendments.

\subsection*{Manual coding}

To date, probably the most fruitful and convincing attempts to capture the amendments introduced by actors and their success in incorporating these amendments into political texts has involved major hand-coding efforts of records pertaining to the political processes of interest.

In a comparative politics context, assessing how actors in different legislatures seek to amend and influence political texts has provided important insights into how the process of legislative review can ameliorate agency problems in political systems with multiparty governments \citep{Martin:2005tr,martin2011parliaments,martin2004policing}. The authors argued that ``given the technical nature of most modern legislation, grasping the policy significance of changes to a draft bill by classifying the substantive content and language of such changes requires extensive expertise in the policy areas dealt with by the bill. [...] Any measure based on our perceptions of substantive policy impact is therefore bound to be highly unreliable, especially when applied to a large number of bills across a variety of policy areas". Instead, they develop ``a more objective measure of change, [defined] as the number of article changes made to the draft version of a government bill" \citep[p.9]{Martin:2005tr}. The data used in these studies was collected through hand coding the changes to legislative texts between the initial proposal and the final piece of legislation decided upon, providing a quantitative, objective and reliable account of how legislation evolved and was amended. In the \citet{martin2011parliaments} study, five distinct Parliaments are considered, with a total of 1,300 legislative proposals across these Parliaments examined. The authors convincingly argue that measuring degrees of change between bills and laws is a good way to examine the inner workings of multi-party governments.

In an international context, the EU has received the most attention in terms of tracking the evolution of political texts, as it is the most well developed international organisation, and holds significant legislative powers. \citet{GEORGETSEBELIS:2001tt} were the first to examine the amendment success of different EU institutions in the legislative process. The authors examined nearly 5,000 separate amendments to a selection of 231 examples of EU legislation negotiated under both co-decision (79) and co-operation (152) between 1988 and 1997. They focused on the exact amendments offered by the actors and the degree to which these are adopted or rejected in the final text. The work by \citet{GEORGETSEBELIS:2001tt} demonstrates that tracking amendments offers a way to objectively measure salient policy developments and the realisation of policy preferences. According to the authors, such tracking is easily quantifiable, non-reliant on subjective experts and undisturbed by dimensionality issues. While advantageous, human coding of amendments is however still time and resource consuming and may be constrained by language issues particularly in cross-national studies. 

\subsection*{Quantitative text analysis}

Building upon the efforts of \citet{GEORGETSEBELIS:2001tt}, \citet{Franchino:2012ie} utilised automated text analysis methods (\textit{Wordfish} \citep{Slapin:2008ud}) developed for ascertaining (ideological) policy positions from political texts, to compare the bargaining success of the Commission, Council, and Parliament in the conciliation committees of the EU. This study is especially useful for the purposes of studying the effectiveness of automated text analysis methods, as the authors make a significant effort to compare the hand-coding scheme used by \citet{GEORGETSEBELIS:2001tt} to the automated method they employ. \citet{Franchino:2012ie} demonstrate that \textit{Wordfish} is capable of producing document-level binary results (i.e. which actor's position is most reflected in the final text) similar to those of hand-coding efforts for a subset (9/20) of the most clear-cut cases, leaving much room for improvement for automated text analysis methods.

One important question that arises when applying automated text analysis to political documents, is whether or not the chosen method is appropriate for the task at hand, given the data-generating processes under consideration. The \textit{Wordfish} algorithm designed by \citet{Slapin:2008ud} aims to place actors responsible for particular political texts on latent ideological policy dimensions, based upon the word frequencies found in the political texts of interest. An important assumption of such an approach made clear by the authors themselves is that the language used in each document can be reduced to a word frequency distribution and that differences between these distributions represent differences between actors on the latent policy dimension being estimated. When one considers \textit{legislative} texts, in which language is highly formalised, this is a rather strong assumption, which may go some way to explaining the rather disappointing success rate the algorithm had in replicating the hand-coding efforts of \citet{Franchino:2012ie}. As argued in Grimmer's (2013) review of the state-of-the-art of automated text analysis, one must pay careful attention to the data-generating process associated with the political texts being analysed in order to be sure that the chosen method is appropriate for task. All of that being said, when such algorithms are applied to the types of texts for which \citet{Slapin:2008ud} originally intended them, the usefulness of automated text analysis become clear. Slapin \& Proksch, and those using their \textit{Wordfish} method appropriately have been very successful in providing important insights into position taking in different parliamentary contexts \citep{Slapin:2010il,Proksch:2009ht,Proksch:2011cz}. 

In research fields outside of political science, other automated methods have been employed to complete tasks similar to that of tracking changes to political texts. In particular, minimum edit distance (MED) algorithms have been developed in bio-informatics, computer science, and natural language processing to measure the number of edit operations (insertions, deletions, or substitutions) required to change one string of characters or words into another \citep{wagner1974order}. They have successfully been applied to problems as diverse as creating accurate spell checkers \citep{wagner1974order,Wagner:1974ut,wong1976bounds}, assessing differences between different dialects in computational linguistics \citep{kessler1995computational,Nerbonne:1997wy}, and assessing genetic alignments in computational biology \citep{Fitch:1967we,Dayhoff:1978vq,Henikoff:1992tk}. The basic structure of the problem of capturing changes between two versions of political text is very similar to the above mentioned applications and minimum edit distance algorithms therefore offer a promising avenue for automated analysis of text amendments.

Hitherto existing minimum edit distance algorithms share one limitation that have hindered their applicability to political texts. They are unable to account for changes involving the transposition of text within a document (commonly called cut-paste changes) wherein one whole section of text is moved from one place to another within a document. Existing MED algorithms record such a change as the entire section being deleted and then rewritten ``from scratch" in the new location, very much over-inflating the degree of change between documents in terms of their contents and the policy implications thereof. 

In applications of text similarity techniques where cut-paste type edits are common (for example plagiarism detection), two types of solutions are commonly observed. Both however have significant drawbacks. The first solution, referred to as fingerprinting, randomly draws sample words from two texts and infers their similarity based on the similarity of the samples \citep{hoad2003methods}. While suitably insensitive to cut-paste reorganisation of text and computationally highly efficient, this technique is rather imprecise and best used to identify possibly similar documents for further similarity checks. The second solution entails the construction of so-called suffix trees, which are essentially indexes of every possible combination of the words in each text. The technique is highly accurate, but constructing suffix trees is computationally very demanding and storing them requires significantly more memory space than storing the texts, making analysis of large bodies of texts impractical \citep[p.39]{smyth2003computing}. When analysing large corpora of text, such computational demands imply that scholars are limited by the hardware that they have access to.

As cut-paste type operations are common in the drafting of political texts, where the order of particular sections of text matter less than the fact that they are included in the document, a new MED algorithm able to efficiently and accurately discern cut-paste operations without over-inflating their semantic and political impact thus has great potential for applications in political science.

With this in mind we now introduce the \textit{Levenshtein} minimum edit distance algorithm, and our own minimum edit distance algorithm referred to as \textit{DocuToads}.\footnote{\textbf{DOCU}ument \textbf{T}ranspose \textbf{O}r \textbf{A}dd, \textbf{D}elete, \textbf{S}ubstitute}

\section*{\textit{Levenshtein} Distance} 

The \textit{Levenshtein} minimum edit distance algorithm is used to assess the differences between two strings of text units, and is calculated as the minimum number of editing operations required to change one string into another \citep{Levenshtein:1966ts}.\footnote{For those interested in a more detailed exposition of the \textit{Levenshtein} distance algorithm we recommend \citet{manning2008introduction}} Three distinct editing operations are allowed by the algorithm, and each has an assigned weight. The allowed editing operations are the deletion of a unit of text (weighted 1), the insertion of a unit of text (weighted 1), or the substitution of a unit of text (weighted 0 if the unit of text does not change and weighted 2 if it does). The algorithm is formalised as follows. $S_{1}(i)$ represents the word in string $X$ at position $i$ and $S_{2}(j)$ is the word at position $j$. The minimum edit distance $D(i, j)$ between two strings, $X = x_{1} \cdots x_{m}$ and $Y = y_{1} \cdots y_{n}$ is the minimum cost of a series of editing operations required to convert text $X$ into text $Y$. The minimum edit distance is computed using a dynamic programming approach, which is a method for solving larger problems by considering a larger problem to be the sum of the solutions to a series of sub-problems \citep{Bellman:1957tx}. A dynamic programming approach allows one to avoid the often high cost associated with recalculating the solution to sub-problems, as such solutions are stored once calculated in a process referred to as memoisation.

\begin{equation}
	D(i, j) = \left\{
	\begin{array}{lll}
		D(i-1, j) + 1,&\\
		[1ex]
		D(i, j-1) + 1,&\\
		[-3pt]
		D(i-1, j-1) &+ &\bigg\{\begin{array}{cc}
			2; & \text{if } S_{1}(i) 	\neq S_{2}(j),\\
			0; & \text{if } S_{1}(i) = S_{2}(j).
		\end{array}
	\end{array} \right.
\end{equation}

As can be seen from the formula, there are three values to be computed at each iteration of the algorithm, and each matrix element $m_{ij}$ corresponds to the minimum of these three values. $D(i-1, j) + 1$ corresponds to a deletion, $D(i, j-1) + 1$ represents an insertion, and $D(i-1, j-1)$ represents a substitution. At each iteration, each element in the matrix is calculated one at a time, taking the values from the previously solved sub-problems as inputs into Formula 1 and solving. In this way the larger problem of converting one string into another is broken down into many distinct individual edit operations, with the minimal path being taken in each iteration.

A second stage of the algorithm allows one to determine the actual edits used to generate the final minimum edit distance score. This is done by starting at position $M_{mn}$, finding which of the three previous possible moves was the least costly, and working backwards through the matrix in this manner. The resulting vector of edit operations is referred to as the backtrace and is useful as it allows one to determine the edit alignment that translates string $X$ into string $Y$.\footnote{It should be noted that there can be more than one path through the matrix that delivers the minimum edit distance, so a backtrace is not necessarily unique.} This can later be used to reconstitute the complete edit history of the documents of interest and determine what has been added, removed, or substituted.

As alluded to above, one weakness of the \textit{Levenshtein} distance is that it not very efficient at accounting for large text transpositions, and places a heavy penalisation on such moves. This is problematic for our application, as such transpositions do not generally imply major policy changes, but instead reflect decisions to change the ordering of provisions within a document. In order to avoid the heavy penalisation associated with moving large sections of a text string using the standard \textit{Levenshtein} distance algorithm, while retaining accuracy and computational efficiency, we have developed a second minimum edit distance algorithm, referred to as \textit{DocuToads} that can account for such transpositions.

\section*{\textit{DocuToads} algorithm} 

The new \textit{DocuToads} algorithm presented here proceeds in two stages, similar to the \textit{Levenshtein} algorithm.\footnote{The code used in this project can be found here: \url{https://github.com/ajhhermansson/DocuToads}} In the first stage, the two texts are reduced to a (sparse) matrix, $\boldsymbol{M}$, in which every cell, $m_{i, j}$, with a non-zero value indicates that a word in the first text,  $S_{1}(i)$, is the same as a word in the second text, $S_{2}(j)$. The exact positive value given to each of those cells when a word match between documents is detected equals the value of the cell above and to the left plus one, $m_{i, j} = m_{i-1, j-1} + 1$ (if $i$ or $j = 1$ and $S_{1}(i) = S_{2}(j)$, $m_{i, j} = 1$). Applying this algorithm iteratively over the entire matrix similar to before creates a matrix with some very useful features for comparing the two texts:

\begin{itemize}
	\itemsep0em
	\item Sequences of matching words are represented in the matrix as diagonal sequences of rising numbers (see Table \ref{tab:toads}(a));
	\item Edit operations are represented by ``gaps" of zeroes interrupting these sequences;
	\item Deletions are represented by a column of zeroes (see Table \ref{tab:toads}(b));
    \item Additions are represented by a row of zeroes (see Table \ref{tab:toads}(c));
	\item Substitutions are represented by a row and a column of zeroes (see Table \ref{tab:toads}(d));
	\item Transpositions are represented by sequences of rising numbers that have been ``shifted" horizontally accompanied by columns of zeroes at a corresponding location (see Table \ref{tab:toads}(e)).
\end{itemize}

\begin{table}
\centering
\caption{\textit{DocuToads} illustration}
\subfloat[Identical text sequences]{
\begin{tabular}{r|cccccc}
\hline
&\rotatebox[origin=c]{270}{A} & \rotatebox[origin=c]{270}{simple} & \rotatebox[origin=c]{270}{minimum} & \rotatebox[origin=c]{270}{edit} & \rotatebox[origin=c]{270}{distance} & \rotatebox[origin=c]{270}{algorithm} \\
\hline
A & \textbf{1} & 0 & 0 & 0 & 0 & 0\\
simple & 0 & \textbf{2} & 0 & 0 & 0 & 0 \\
minimum & 0 & 0 & \textbf{3} & 0& 0 & 0 \\
edit & 0 & 0 & 0 & \textbf{4} & 0 & 0\\
distance & 0 & 0 & 0 & 0 & \textbf{5} & 0\\
algorithm & 0 & 0 & 0 & 0 & 0 & \textbf{6}\\
\hline
\end{tabular}
}

\subfloat[Deletion]{
\begin{tabular}{r|cccccc}
\hline
&\rotatebox[origin=c]{270}{A} & \rotatebox[origin=c]{270}{\textbf{simple}} & \rotatebox[origin=c]{270}{minimum} & \rotatebox[origin=c]{270}{edit} & \rotatebox[origin=c]{270}{distance} & \rotatebox[origin=c]{270}{algorithm} \\
\hline
A & 1 & \textbf{0} & 0 & 0 & 0 & 0\\
minimum & 0 & \textbf{0} & 1 & 0& 0 & 0 \\
edit & 0 & \textbf{0} & 0 & 2 & 0 & 0\\
distance & 0 & \textbf{0} & 0 & 0 & 3 & 0\\
algorithm & 0 & \textbf{0} & 0 & 0 & 0 & 4\\
\hline
\end{tabular}
}
\subfloat[Addition]{
\begin{tabular}{r|cccccc}
\hline
&\rotatebox[origin=c]{270}{A} & \rotatebox[origin=c]{270}{simple} & \rotatebox[origin=c]{270}{minimum} & \rotatebox[origin=c]{270}{edit} & \rotatebox[origin=c]{270}{distance} & \rotatebox[origin=c]{270}{algorithm} \\
\hline
A & 1 & 0 & 0 & 0 & 0 & 0\\
simple & 0 & 2 & 0 & 0 & 0 & 0 \\
\textbf{new} & \textbf{0} & \textbf{0} & \textbf{0} & \textbf{0} & \textbf{0} & \textbf{0} \\
minimum & 0 & 0 & 1 & 0& 0 & 0 \\
edit & 0 & 0 & 0 & 2 & 0 & 0\\
distance & 0 & 0 & 0 & 0 & 3 & 0\\
algorithm & 0 & 0 & 0 & 0 & 0 & 4\\
\hline
\end{tabular}
}

\subfloat[Substitution]{
\begin{tabular}{r|cccccc}
\hline
&\rotatebox[origin=c]{270}{A} & \rotatebox[origin=c]{270}{\textbf{simple}} & \rotatebox[origin=c]{270}{minimum} & \rotatebox[origin=c]{270}{edit} & \rotatebox[origin=c]{270}{distance} & \rotatebox[origin=c]{270}{algorithm} \\
\hline
A & 1 & \textbf{0} & 0 & 0 & 0 & 0\\
\textbf{new} & \textbf{0} & \textbf{0} & \textbf{0} & \textbf{0} & \textbf{0} & \textbf{0} \\
minimum & 0 & \textbf{0} & 1 & 0& 0 & 0 \\
edit & 0 & \textbf{0} & 0 & 2 & 0 & 0\\
distance & 0 & \textbf{0} & 0 & 0 & 3 & 0\\
algorithm & 0 & \textbf{0} & 0 & 0 & 0 & 4\\
\hline
\end{tabular}
}
\subfloat[Transposition]{ 
\begin{tabular}{r|cccccc}
\hline
&\rotatebox[origin=c]{270}{\textbf{A}} & \rotatebox[origin=c]{270}{\textbf{simple}} & \rotatebox[origin=c]{270}{minimum} & \rotatebox[origin=c]{270}{edit} & \rotatebox[origin=c]{270}{distance} & \rotatebox[origin=c]{270}{algorithm} \\
\hline
minimum & 0 & 0 & 1 & 0& 0 & 0 \\
edit & 0 & 0 & 0 & 2 & 0 & 0\\
distance & 0 & 0 & 0 & 0 & 3 & 0\\
algorithm & 0 & 0 & 0 & 0 & 0 & 4\\
\textbf{A} & \textbf{1} & 0 & 0 & 0 & 0 & 0\\
\textbf{simple} & 0 & \textbf{2} & 0 & 0 & 0 & 0 \\
\hline
\end{tabular}
}
\label{tab:toads}
\end{table}

It is possible to draw paths from the cell representing the start of both texts, $m_{1, 1}$, to the cell representing the end of both texts, $m_{I, J}$. Each such path represents a possible edit history, i.e. a way in which $S_{1}$ could have been transformed into $S_{2}$. In the case of two identical texts (see Table 1(a)) the path that involves the least edits would simply be to leave the first text as it stands (following the path set by the rising numbers along the matrix diagonal), but it would also be possible to delete all the words and rewrite them again (following a path vertically down from $m_{1, 1}$ to $m_{1, J}$ and then horizontally to $m_{I, J}$). For any two texts, there thus exists a very large number of potential edit paths, with the edit path that reflects the actual edit procedure used to create a document being the most substantively interesting. The second stage of the algorithm is designed to find the specific edit path from $m_{I, J}$ upwards through the matrix, row by row, to $m_{1,1}$ that involves the fewest edit operations.

Before moving on to describing that second stage, it should be noted that there are limits to which edit paths represent a valid edit history (there is always at least one valid path). First, any edit path that does not pass through every row and every column is invalid, since that implies the full texts have not been considered. Furthermore, edit paths that pass more than one cell per row or column are also invalid as they imply copy-paste (not cut-paste) edit operations. We argue that copy-paste edit operations are not a realistic way of describing the drafting of legislative text, in which pure repetition would be superfluous. If a path, $D_{i,j,p}$, is described as a three-dimensional vector where indices i and j indicate a position in $M_{i, j}$ and a step on the path is described as $d_{i,j,p}$ and $d_{p}(i,j) = m_{i,j}$. The above two rules for detecting a valid edit path can then be expressed as (i) $D_{i,j,p}$ is valid when $D_{i}(p>0) \subseteq \left\{1:I\right\}$ and $D_{j}(p>0) \subseteq \left\{1:J\right\}$ or (ii) $d_{i,j,p}$ is valid when $d_{i,j}(p>0)$ is not in $D_{i}(p>0)$ or $D_{j}(p>0)$. The set of steps $d_{i,j}$ that are allowed under the above rules are termed $Q_{i,j}$.

Finding the edit path with the fewest edit operations that satisfies the above rules is a problem that is once again solvable using dynamic-programming methods, i.e. by breaking the problem down into the sub-problem(s) of choosing what the next step of the path should be.\footnote{Another type of solution might be to iterate over all possible paths and choose that which produces the lowest MED, but this would be exponentially more computationally demanding.} As with the \textit{Levenshtein} algorithm, this is achieved by assigning different penalties to different steps and having the algorithm choose the step with the lowest associated penalty. These penalties must be assigned in such a way that following the locally lowest penalty produces the least costly overall path.  In the \textit{DocuToads} algorithm, the possible steps $d_{k,l}$ from starting point $m_{i,j}$ to destination $m_{k,l}$ and their respective penalties and conditions are detailed in Table \ref{tab:toadsteps}.\footnote{Different MED algorithms feature different sets of penalties, as illustrated by comparing these penalties to those of the \textit{Levenshtein} algorithm.}

\begin{table}
\centering
\caption{\textit{DocuToads} steps and penalties}
\begin{tabular}{l|l|l|l}
\hline
\textbf{Step} & \textbf{Penalty} & \textbf{Destination} &	\textbf{Condition} \\
\hline
No edit operation &	0	& $m_{i-1, j-1}$	& $m_{i-1, j-1} = max(m_{i-1}) > 0$ \\\hline
\multirow{2}{*}{Transposition} & \multirow{2}{*}{1} &	\multirow{2}{*}{$m_{i-1, l}$} &	a) $m_{i-1, l} = max(Q_{p}(i-1)) = max(m_{l}) > 5$\\
& & & b) $|l-j| = min(\{|Q_{l}(i-1,p=m_{i-1,l})-j|\})$\\\hline
Substitution &	1	& $m_{i-1, j-1}$ &	$max(Q_{i-1}) = max(Q_{j-1})  =0$ \\\hline
Addition &	2	& $m_{i-1, j}$ &	$max(Q_{i-1}) = 0$ \\\hline
Deletion &	2 &	$m_{i, j-1}$ &	$max(Q_{j-1}) = 0$ \\\hline
Any other step &	3	& $m_{k,l}$	&  None \\\hline\hline
\end{tabular}
\label{tab:toadsteps}
\end{table}

Setting aside the highest penalty for the moment, the least efficient types of steps are additions and deletions. This is because substitutions, when possible, replace one of each. The conditions under which additions and deletions are allowed never coincide while the conditions for a more efficient substitution are unfulfilled. Transpositions, when a large enough section of text has been moved, also represent a more efficient edit operation than additions and deletions.\footnote{Technically, the shortest but still efficient transposition involves moving two words. However, to avoid finding spurious cut-paste operations when a more likely cause is the recurrence of common words such as ``the", ``a", etc., the shortest transposition length to be considered is set at 5. This can be adjusted as desired, although a trade-off between avoiding finding spurious cut-paste operations and missing short but non-spurious matching sequences exist.} Moving for example $100$ words to a different position in a text would either require deleting $100$ words and adding $100$ words at the new location or one cut-paste operation. These two operations cannot be directly compared in terms of efficiency because there is no way to replace a transposition with a substitution. They therefore share the same penalty in the algorithm, but the conditions under which they are possible never coincide. If it is possible to turn one string of text into another without editing it, performing no edit operations is trivially more efficient than performing edit operations and so carries the lowest possible penalty. 

Since we are uninterested in copy-paste operations (because of their assumed irrelevance and non-application in legislative drafting where large repeated sections of text would serve little function) we have to consider that a word in one text can only be matched with a single word in the other text. In particular, the algorithmic rules must prevent steps involving non-edits or transpositions from precluding other longer matching sequences. The highest penalty (corresponding to the last row of Table 2) is therefore given to these operations when $m_{l,k} < max(m_{k})$, because taking such a step will invariably cause at least one edit operation further down the path. Excluding this rule would mean prioritising shorter matching sequences towards the end of texts above longer sequences earlier in texts, but would still produce edit paths with the same number of edit operations. The transposition conditions a) $m_{i-1, l} = max(Q_{p}(i-1))$ and b) $|l-j| = min(\{|Q_{l}(i-1,p=m_{i-1,l})-j|\})$ ensure that only the best transpositions for each row are considered (the closest highest target value).

As \textit{DocuToads} applies these rules from a starting point at $m_{I,J}$ it generates a backtrace through the matrix similar to the \textit{Levenshtein} algorithm. In practice as both algorithms are applied, we store each step, it's coordinates in the matrix, the two words from their respective texts, and the type of (non-)edit operation that occurs at each algorithm iteration. These data can subsequently be used t o complete a number of very useful operations that canprovide substantive insight into how texts have evolved, including:

\begin{enumerate}
	\itemsep0em
	\item \textit{Reconstruction} of one text from the other in order to see how a document evolved;
    \item Sorting of edit operations into articles (or paragraphs, sections or any other sub-unit of the texts); for \textit{article-level analysis} (with the help of regular expressions to detect article breaks)
    \item \textit{Extraction} of each change (or similarity) and its \textit{context} within a particular document;
    \item Production of a total count of the \textit{number} and \textit{types} of each (non-)edit operation.
\end{enumerate}

These different operations provide huge scope for exploring how draft legislation and other types of political texts evolve, and for demonstrating how different actors have influenced the legislative process. Table \ref{tab:backtrace} demonstrates the format of the backtrace produced by \textit{DocuToads}.\footnote{The Levenshtein algorithm can produce a backtrace of the same format.}

\begin{table}
\centering
\caption{Illustration of backtrace}
\begin{tabular}{c|c|c|c|c}
\textbf{$S_{1}$ index} & \textbf{$S_{2}$ index} & \textbf{Edit operation} & \textbf{$S_{1}$ word} & \textbf{$s_{2}$ word}\\
\hline
375 & 440 & & Whereas & Whereas\\
376 & 441 & &the& the\\
377 & 442 & &European &European\\
378 & 443 & &Community & Community\\
379 & 444 & Substitution & can & is\\
379 & 445 & Addition & &in\\
379 & 446 & Addition & &a\\
379 & 447 & Addition & &position\\
379 & 448 & Addition & &to\\
380 & 449 & & make & make\\
381 & 450 & & a& a\\
381 & 451 & Addition& & major\\
382 & 452 & & contribution & contribution\\
383 & 453 & & towards& towards\\
384 & 454 & & the & the\\
385 & 455 & & organisation& organisation\\
386 & 456 & & of & of
\label{tab:backtrace}
\end{tabular}
\end{table}

Users of minimum edit distance algorithms can observe the semantic context of changes by extracting a section of the backtrace immediately around the edit operations of interest, as demonstrated in table \ref{tab:backtrace}.\footnote{This type of output is then amenable for further analysis.} For well formatted text, it is possible to extract the specific sentence(s) in which changes occurred. It is also possible to draw on word frequency or dictionary methods to weight the changes found therein. The extensive output produced by minimum edit distance algorithms of course also facilitates human reliability checks. 

It should be well noted that minimum edit distance algorithms can identify the most efficient way of turning one text into another, in terms of the number of needed edit operations. They do not and cannot however identify which edit operations that were actually performed by the human editor to affect the observed change. For example, while it may be most efficient to substitute a section of text, the editor may in fact have deleted a section of text and rewritten a new section on a different topic. Transpositions in particular often offer a very efficient way of transforming a text, more efficient than actual human use. The backtrace produced by MED algorithms and a (hypothetical) log of actual changes performed will therefore not match completely. With this caution in mind, we now move on to compare the performance of MED algorithms with other existing methods.

\section*{Comparison with existing measures}

In this section we show that our method can replicate and improve upon the results of hand-coded analyses found in the existing literature. Since legislative texts represent an ideal case for our method and simultaneously have received intensive scholarly attention, we focus on replicating studies from this field. We shall first replicate the coding of substantively important legislative amendments to secondary European law performed by \citet{Franchino:2012ie}. We aim to demonstrate that automated methods can perform just as well as and sometimes better than hand-coding efforts, regardless of the chosen algorithm, and that our new \textit{DocuToads} algorithm outperforms the existing \textit{Levenshtein} algorithm due to its ability to detect transpositions in legislative texts. We then move on to compare the \textit{DocuToads} algorithm to a sample of the hand-coding performed by \citet{Martin:2005tr}, aiming to replicate their results and to compare the counting of article changes to counting edit operations.

\subsection*{The Conciliation Committee of the EU}

In the EU, most legislation is enacted through the ordinary legislative procedure (formerly known as the codecision procedure) in which power is shared between the European Commission, which puts forward legislative proposals, and the European Parliament and the Council of Ministers, who can amend those proposals. For legislation to be enacted, majorities of both the Parliament and the Council have to agree on a particular version of the law. If a compromise text cannot be agreed upon after two rounds of negotiation within and between the institutions, the proposal is referred to a Conciliation Committee where representatives of all three institutions negotiate a joint compromise text or the law finally fails (if the joint text is voted down by either the Council or Parliament). This decision-making mechanism has become an important institutional feature of the EU \citep{Farrell:2004cu}.

Essentially, the different versions of a piece of legislation put forward by the Council and Parliament reveal overt conflict between the central institutions and the final text drafted by the Conciliation Committee represents the resolution of that conflict. Taking the Council of Minister's version as a reference document, \citet{Franchino:2012ie} have encoded substantive differences between the three versions of twenty legislative acts that were resolved by Conciliation Committee on an article-by-article basis, using the same methodology and terminology as \citet{GEORGETSEBELIS:2001tt}. All-in-all, the authors categorised $525$ substantive amendments in the recitals and articles of the texts and associated them with particular legislative articles.\footnote{They also coded amendments in the annexes and appendices, but we ignore these as the formatting is too irregular to identify the same subsections as Franchino and Mariotto, since their cleaning of the texts removed subsection names.}

\begin{figure}[ht!]
\centering
\caption{Document-level relationship between Franchino and Mariotto coding and \textit{DocuToads}}
\includegraphics[width = 10cm]{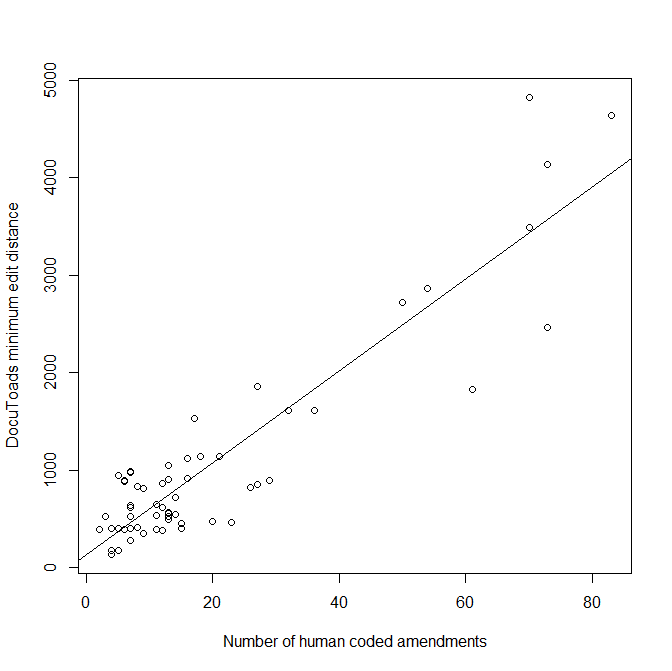}
\label{fig:DTvFM}
\end{figure}

In order to demonstrate that minimum edit distance measures are capable of replicating these hand-coding efforts, we show the similarity of the \textit{Levenshtein} and \textit{DocuToads} minimum edit distances to this hand-coded scheme in two ways. First, we take each document-level dyad (e.g. Council-Parliament, Council-final text, Parliament-final text) as the unit of analysis, resulting in $60$ document combinations to analyse ($20$ dossiers with three document combinations in each). For each such dyad, we add up the number of amendments found by \citet{Franchino:2012ie} and compare what they find to the \textit{Levenshtein} and \textit{DocuToads} minimum edit distances between the documents of interest. While the \textit{Levenshtein} edit distance and human coding are highly correlated, with an R-squared of $0.73$ in a simple linear model, \textit{DocuToads} shows an even higher correlation with the hand-coded measure, comparison yields an R-squared of $0.84$. The relationship between the \textit{DocuToads} edit distance and the Franchino \& Mariotto measure is demonstrated in Figure \ref{fig:DTvFM}. Furthermore, when comparing the two minimum edit distance algorithms, we find that they are strongly correlated with an R-squared of $0.93$. This suggests that both algorithms do a good job of replicating the efforts of the hand coders.

\begin{figure}[ht!]
\centering
\caption{Article-level relationship between Franchino and Mariotto coding and \textit{DocuToads}}
\includegraphics[width = 10cm]{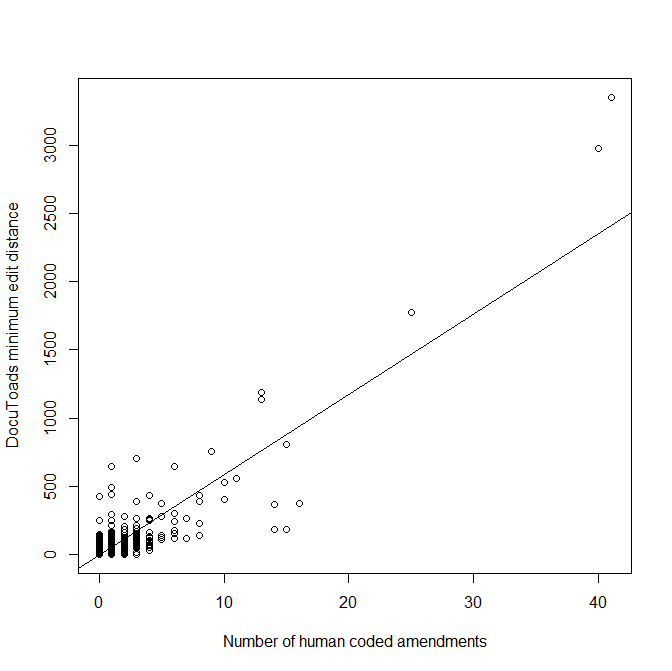}
\label{fig:DTvFM_art}
\end{figure}

Second, we add up the amendments on an article-by-article basis.\footnote{The Franchino and Mariotto coding scheme allows for multiple amendments to the same article.} We then compare that number of amendments to the article-specific minimum edit distances produced by \textit{DocuToads} and the \textit{Levenshtein} algorithm. We do this for all three sets of document dyads. The unit of analysis is thus an article in one of the texts.\footnote{The reference document is the Council's version, except in the comparison between the Parliament and final text, when the parliamentary version is taken as the reference document.} There are in total $2,328$ such articles. Once again, while the \textit{Levenshtein} minimum edit distance and hand-coding are highly correlated (R-squared of $0.70$), \textit{DocuToads} and hand-coding are more so (R-squared of $0.76$) in a simple linear model. The article-level relationship between the \textit{DocuToads} results and hand-coding of amendments is demonstrated in Figure \ref{fig:DTvFM_art}. Comparison of the two minimum edit distances yielded an R-squared of $0.94$. These correlations are a large step forward compared to the performance of the \textit{Wordfish} algorithm \textit{in this setting}, which could only provide document-level results correctly indicating in a binary fashion whether the Parliament or Council was closest to the final text in nine out of twenty cases. We emphasise that the poor performance of the \textit{Wordfish} algorithm is a result of its application to phenomena in which the data generating processes are very different from those for which the method was developed.

Even though the results presented above are highly encouraging in terms of the validity of minimum edit distances in general and \textit{DocuToads} in particular for replicating hand-coding of substantive amendments, we shall now further explore the differences that do exist between the measures, as this can inform us about the advantages and disadvantages of using the proposed method. To do so, we focus on type-I (false positive) and type-II (false negative) errors in order to examine when each method under cosideration fails to account for substantive changes to a legislative text. We examine the $100$ articles where \textit{DocuToads} and the hand-coding by Franchino and Mariotto disagree the most, in terms of the fifty largest positive and negative residuals, respectively.\footnote{We focus on the \textit{DocuToads} errors given the high correlation between both algorithms, and the fact that it performs best in replicating hand-coding efforts.} By studying these $100$ legislative articles (and, when necessary, the surrounding context), the \textit{DocuToads} output, and the hand-coding notes, we classified the following main sources of disagreement: 

\begin{itemize}
	\itemsep0em
  \item \textit{DocuToads} and human coders identify the same changes but place them in adjacent articles ($39/100$ articles);\footnote{Removing these articles reduced the residual variance by $29\%$}
  \item Small changes in article language still counted as an amendment by human coders ($31/100$ articles);\footnote{Removing these articles reduced the residual variance by $1\%$}
  \item \textit{DocuToads} identifies substantive changes that human coders missed ($15/100$ articles);\footnote{Removing these articles reduced the residual variance by $10\%$}
  \item Large changes in article language counted as only one amendment by human coders ($11/100$ articles);\footnote{Removing these articles reduced the residual variance by $25\%$}
  \item \textit{DocuToads} detects substitutions where human coders identify added and removed sections of text as two distinct amendments ($4/100$ articles).\footnote{Removing these articles reduced the residual variance by $1\%$}
\end{itemize}

The most common source of disagreement in our sample of articles can be traced primarily to the addition of new articles. In such cases, new text was added between two previously existing articles in the reference document. \textit{DocuToads}, however, has to record the additions at an index position in the reference text which will correspond to either the preceding or following article. Depending on the edit operations performed inside those two articles, it will be less costly to place the added text at the end of the preceding or beginning of the following article (see Table \ref{tab:toads}). There is in other words a random component which may cause \textit{DocuToads} to disagree with human coders who can implement any single rule on the placement of new articles.\footnote{\citet{Franchino:2012ie}
did not indicate where in the reference document an added article belongs, we therefore placed such articles as belonging to the preceding article when implementing their coding system.} This constitutes a form of negative serial correlation, a common concern found in time-series analysis. Users of \textit{DocuToads} are advised to consider this issue if correlating article-level results with other article-level variables assigned by human coders.\footnote{A potential solution is to use regular expressions to detect and index the end of articles, thereby creating the possibility to place edit operations between articles. This solution was not implemented for this set of documents, as it is a side issue unrelated to the core of the method.} This source of disagreement has no effect on the document-level results. 

The second and fourth most common sources of disagreement concern the length vis-a-vis substantive importance of individual amendments. The underlying assumption we make when interpreting the \textit{DocuToads} output is that larger amendments are likely to be more substantively important. This assumption is justified to a large extent by the results presented above, but there are certain language features that still require attention. In particular, negations and altered numbers (setting for example a budget) can cause large semantic changes with small alterations to the text. In such cases, our assumption about the size of an amendment corresponding to its importance is challenged, resulting in a false-negative type-II error or an under-estimation of the substantive importance of a amendment. 

On the other hand, the amendments coded by \citet{Franchino:2012ie} are binary and thus conceal variation in the importance and size of amendments, which \textit{DocuToads} arguably captures more accurately. When hand-coders fail to account for the correspondence between the size and importance of an amendment, false-negative type-II errors or an under-estimation of the importance of an amendment can arise. There are in other words arguments both in favour of the automated and human-coded description of these amendments. 

Even though it wasn't a major driver of the differences between \textit{DocuToads} results and human coding, another potential source of disagreement could be that \textit{DocuToads} picks up unsubstantive language changes or spelling errors that human coders ignore. Some of these spelling errors could be the result of the pre-processing of documents necessary to run the automated text analysis. In particular, care must be taken when converting PDF documents to pure text as machine-reading errors, column and footnote placement, as well as certain copy-protections may interfere. These obstacles can generally be overcome but do require thorough consideration.\footnote{Remember `garbage in, garbage out'.} Applying spell-checking software on the texts can sometimes be advantageous to reduce noise.

The third most common source of disagreement is that \textit{DocuToads} identified substantively important changes to the text that human coders missed. This represents another form of type-II error on the part of the human coders. In these cases, our re-reading of the articles confirmed the substantive changes identified by \textit{DocuToads}. Finally, in four of the articles we examined, articles with a high number of amendments made to them, the main cause of disagreement was that \textit{DocuToads} recorded changes as substituted text while human coders recognised that the removed text was on a different topic than the added text and that it was more appropriate to record the change as two distinct amendments rather than one substituted section of text. This type of error can be thought of as a type-II error on the part of the \textit{DocuToads} algorithm, as two substantively important amendments are counted as one resulting in an under-estimation of the true amount of substantive change involved.

As mentioned above, the two minimum edit distance algorithms are, as is to be expected, very strongly correlated. Out of the sixty document dyads we examined, there were however four cases for which there was substantial disagreement between the two measures, and which caused the \textit{Levenshtein} algorithm to disagree more with the hand-coded measure. These four cases involved two documents. The first of these was the Council of Ministers' version of the 1998 act establishing the EU educational programme Socrates. Specifically, \textit{DocuToads} recorded 396 and 972 edit operations between the Council version and Parliament version and joint text, respectively, while the \textit{Levenshtein} algorithm recorded 1,768 and 1,695 edit operations. The second document was the joint text version of the 1995 act on government procurement from third countries, in relation to which the \textit{Levenshtein} algorithm recorded 1,460 and 1,479 changes while \textit{DocuToads} recorded 893 and 887 changes, respectively. In these cases the \textit{Levenshtein} algorithm in other words recorded between 1.7 and 4.4 times more edit operations than \textit{DocuToads}. Comparison of the backtraces of both algorithms reveals that these are cases in which transposition of large sections of text played an important role. For example, the list of possible Socrates actions was moved from Article 2 to Article 3 while the transitional measures were moved further down within Article 5 and preambles concerning the system for transfer of European university credits were reshuffled. The \textit{Levenshtein} algorithm, unable to correct for transpositions, commits a type-I error by recording these examples as first deleted and later re-added sections of the text, overestimating the substantive importance of these rather cosmetic changes that had been ignored by hand coders. In conclusion, \textit{DocuToads} is to be preferred in cases where cut-paste operations are common or expected while the \textit{Levenshtein} algorithm is slightly faster and equivalent in the absence of such operations. In cases where no transpositions but overall high numbers of edit operations are to be expected, the Levenshtein algorithm will also provide a backtrace more closely corresponding to a (hypothetical) log of actual changes performed by the writer.

In conclusion to this section, we are very confident in the ability of minimum edit distance algorithms in general and \textit{DocuToads} in particular to replicate and replace hand coding for the coding of substantive amendments to legislative text. Our results are highly consistent with hand coding, and much more consistent than the automated method employed by \citet{Franchino:2012ie}. The differences that do exist are almost as likely to be the result of human error as of the failures of minimum edit distance algorithms. Given the fact that minimum edit distances are transparent and very reliable in how they operate, and that they produce the same result every time without the inter-coder sources of error common to all hand-coding efforts, minimum edit distance algorithms are in many cases superior to hand coding.

\subsection*{Parliamentary amendments in Germany}

\begin{figure}[ht!]
\centering
\caption{Relationship between Martin and Vanberg coding and number of articles with \textit{DocuToads} MED $>$ 25}
\includegraphics[width = 10cm]{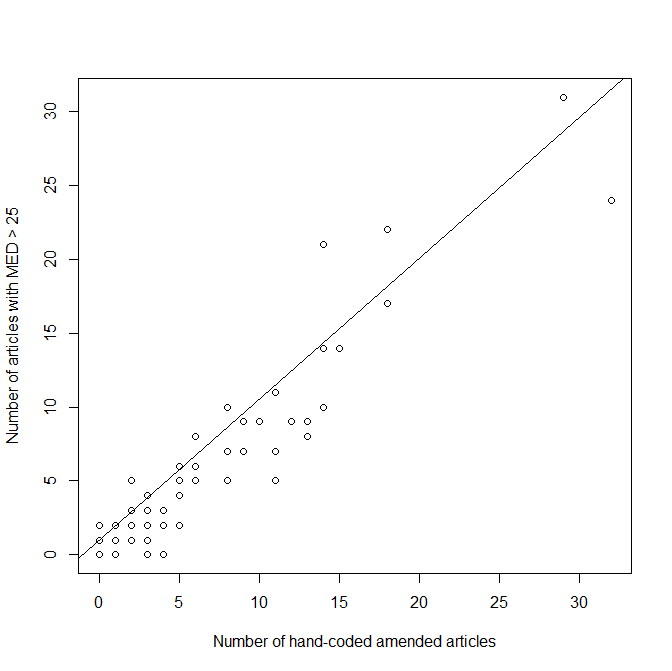}
\label{fig:DT25vVM}
\end{figure}

In a series of valuable contributions to the literature on multiparty governments, Martin and Vanberg \citep{Martin:2005tr,martin2011parliaments,martin2004policing} developed a quantitative measure of the amount of change between draft bills proposed by governments and final laws passed by parliamentary bodies. Arguing that each article of a bill represents a distinct aspect of the proposed legislation, the authors counted the number of articles subjected to change, and produced a document-level measure of amendment success. While the authors did not provide us with the original texts they had worked with, we were able to identify and attain 66 out of 148 document pairs from their work on the German Bundestag \citep{Martin:2005tr}. In a first step towards replicating their measure, we ran the \textit{DocuToads} algorithm on the document pairs (bill-law) and then sorted the changes into the articles in which they happened (using a regular expression to delineate the articles). We then calculated the number of articles subjected to more than 25 edit operations as a threshold for identifying amended articles to allow for minor language changes that do not represent substantively important amendments.\footnote{We optimised the cutoff point to achieve the best fit, but this threshold is easily adjustable.} The results are presented in Figure \ref{fig:DT25vVM} and are highly encouraging. A linear regression of the two measures showed a high degree of correlation (R-squared is $0.87$) between the automated and hand-coded measure. Close reading of the bills and laws revealed somewhat different standards regarding the way in which other pieces of legislation was referred to; bills tended to cite while laws tended to reference with paragraph numbers. This introduced some error to the \textit{DocuToads} results which accounted for a large part of the disagreement between the measures and can be understood as a difference in the data-generating processes of the two document types.

\begin{figure}[ht!]
\centering
\caption{Relationship between Martin and Vanberg coding and \textit{DocuToads} MED}
\includegraphics[width = 10cm]{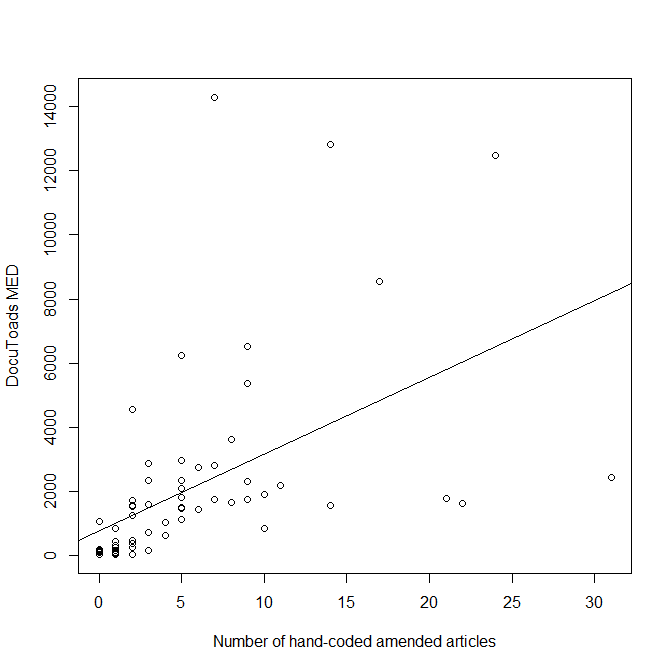}
\label{fig:DTvVM}
\end{figure}

An interesting aspect of our analysis is that it revealed significant variation in the amount of changes necessary for Martin and Vanberg to code an article as amended. For some document pairs, the average number of amended words needed for a change to be recorded by human coders was as low as three. For other documents, especially those with long articles, that number was in the order of hundreds. This reflects the difficulty for human coders to reliably implement a binary measure of change at the article level, and serves to highlight the usefulness of a continuous and fully replicable automated measure.\footnote{Binary on the article level, a count variable on the document level.} We similarly found that the number of articles changed between bill and law, in contrast to the \citet{Franchino:2012ie} measure which could detect several substantive amendments within each article, was unable to convey the extent of changes an article had been subjected to. Some articles had received amendments in the order of several thousand words but were naturally only coded as a single amendment at the article level. Whether this fact is problematic depends on whether we believe a thousand word amendment is more substantively important, in terms of policy consequences or political capital, than a very short amendment. Figure \ref{fig:DTvVM} illustrates the correlation between the number of amended words as captured by \textit{DocuToads} and number of amended articles as captured by hand coding (R-squared is $0.26$). 

The average document length in this sample was just over 26,000 words and the average time to process one document pair was just under half a minute on a standard desktop computer using \textit{DocuToads}, highlighting the great gain in efficiency possible compared to hand-coding.\footnote{Both minimum edit distances were implemented using the Python programming language. This is a powerful and fast language that is designed to interact well with other applications, features useful modules for processing text and data in all formats, runs on any operating system and is free.} Parallel processing on several computing cores simultaneously is available and further greatly reduces the time needed.

\section*{Conclusions}

In this study we have demonstrated the usefulness of minimum edit distance algorithms for quantifying differences between versions of political texts. We demonstrate that these algorithms are particularly suitable to detecting amendments or changes to legislation. Furthermore, we introduced a new minimum edit distance algorithm which can handle text transpositions, i.e. cut-paste operations, in a way that previously existing MED algorithms could not. 

In the empirical section of the study we compared the results of the \textit{DocuToads} algorithm and the previously existing \textit{Levenshtein} algorithm with the substantive amendments identified and hand-coded by \citet{Franchino:2012ie}, who followed the same procedure as \citet{GEORGETSEBELIS:2001tt}. We found a strong correlation between both MED results and the hand-coded measure on the document as well as article level. The two algorithms performed equally well except for the set of cases where transpositions played an important role. In such cases, the suitability of the \textit{DocuToads} algorithm in particular was demonstrated due to its ability to account for transpositions and avoid the type-I errors that the \textit{Levenshtein} algorithm is vulnerable to. Furthermore, we compared the \textit{DocuToads} algorithm to the hand-coded scheme employed by \citet{Martin:2005tr,martin2011parliaments,martin2004policing}. Our method proved capable of replicating their measure with a very high degree of accuracy while also providing more detail and overcoming some unavoidable difficulties associated with hand-coded binary measurements. We also demonstrated the speed of the new method. The results presented in this study suggest that our proposed method is capable of capturing changes to political texts in a very fine-grained and replicable manner.

Bearing these strengths in mind, the method naturally also has some limitations. We have highlighted that users must carefully consider first of all the purpose of study and precisely what the size of amendments between text versions signifies in the context of interest. Second, it is supremely important to consider the data-generating process underlying the production of texts. In particular, when applying minimum edit distances, it is important that the texts are redrafted rather than rewritten, and that every word matters, both substantively and politically. Finally we want to emphasize that as with all automated text analysis methods the principle of `garbage in garbage out' applies. Users should ensure a consistent and error-minimizing pre-processing of documents to achieve best results.

The new methodology applied in this study has huge potential to inform us about the manner in which political documents are produced, and the influence that different actors have over their drafting. In situations where the data-generating process resembles an iterative adaptation of a political text, and in which the actor associated with each iteration can be identified, our method can produce a fine-grained measure of the success each actor had in influencing the final document. Applications that come to mind include further investigations of legislative decision making from a comparative politics perspective, drafting of international treaties, routine bureaucratic decisions and the drafting of party manifestos. This study should thus act as a catalyst for further explorations into the usefulness of these methods to investigate the drafting of political texts.

\bibliographystyle{apalike}
\bibliography{Bib_lib}

\end{document}